\documentclass[]{fairmeta}
\title{QQWorld: Quantile-Quantile Matching for World Model Regularization}

\author[]{Zhoushun Yu}
\author[*]{Xiaoyu Hu}
\author[*]{Xiangyu Xu}

\affiliation{Xi'an Jiaotong University}

\contribution[*]{Corresponding author}

\newcommand{\method}{QQWorld}

\abstract{
\vspace{-5pt}
Latent world models enable efficient planning by predicting future states in a compact representation space, but their performance depends critically on the quality of the learned latent distribution. LeWorldModel (LeWM) regularizes its latents toward an isotropic Gaussian using the Epps--Pulley (EP) objective. We show that the corrective gradients of EP rapidly vanish for isolated tail samples, leaving heavy-tailed deviations insufficiently controlled. To address this limitation, we propose \emph{QQWorld}, which replaces EP with a quantile--quantile matching objective that directly aligns projected latent samples with rank-matched Gaussian quantiles, thereby maintaining effective corrective gradients in the tails. We further develop cross-batch QQ, which enlarges the effective ranking pool using detached samples from previous batches, and characterize its bias--variance trade-off. Across four control environments, QQWorld effectively improves the average planning success rate of LeWM, while consistently yielding better Gaussian alignment and thinner latent tails. 
\vspace{-8pt}
}

\usepackage{hyperref}
\usepackage{url}
\usepackage[T1]{fontenc}

\usepackage{booktabs}
\usepackage{amsfonts}
\usepackage{nicefrac}
\usepackage{microtype}

\usepackage{epsfig, xspace, enumitem}

\usepackage{
  graphicx,
  amsmath,
  amssymb,
  caption,
  subcaption,
  multirow,
  overpic,
  textpos,
  pifont,
  adjustbox
}

\usepackage{xcolor}
\usepackage[utf8x]{inputenc}

\makeatletter
\@namedef{ver@everyshi.sty}{}
\makeatother

\usepackage{pgf, tikz, pgfplots}
\usepackage{makecell}
\usepackage{pgf-pie}
\usepackage{float}
\usepackage{wrapfig}
\usepackage{colortbl}

\usepackage{newfloat}
\usepackage{listings}
\usepackage{siunitx}
\usepackage{amsthm}

\usepackage{comment}
\usepackage{threeparttable}

\DeclareUnicodeCharacter{3010}{[}
\DeclareUnicodeCharacter{3011}{]}

\newtheorem{proposition}{Proposition}

\theoremstyle{definition}

\theoremstyle{remark}

\newcommand{\missingfigure}[1]{%
  \fbox{%
    \parbox[c][0.32\textheight][c]{0.94\linewidth}{%
      \centering
      Figure file missing: #1%
    }%
  }%
}

\newcommand{\paperfig}[2][]{%
  \IfFileExists{figs/#2}{%
    \includegraphics[#1]{figs/#2}%
  }{%
    \missingfigure{#2}%
  }%
}

\setlist[enumerate]{
  itemsep=-0.5mm,
  partopsep=0pt
}

\pgfplotsset{compat=1.18}

\begin{document}

\maketitle

\section{Introduction}
Latent world models aim to learn a latent space in which an agent can predict dynamics and plan~\citep{leworldmodel2026,fastlewm2026}. An important design choice in recent latent world models is to regularize the marginal distribution of the learned latents toward an isotropic Gaussian, which has been shown to uniquely minimize downstream prediction risk~\citep{balestriero2025lejepa}.

Rather than matching only low-order moments~\citep{bardes2022vicreg}, a growing line of work enforces this constraint with a full normality test as a differentiable penalty, which supplies a distribution-level training signal. In particular, LeWorldModel (LeWM)~\citep{leworldmodel2026} adopts the Epps–Pulley (EP) test, a classical normality test built on the characteristic function~\citep{epps1983test}, as the distributional regularizer for latent world models.

\begin{figure*}[!t]
\centering
\includegraphics[
width=0.93\textwidth,
keepaspectratio, trim={0 6mm 0 0},clip
]{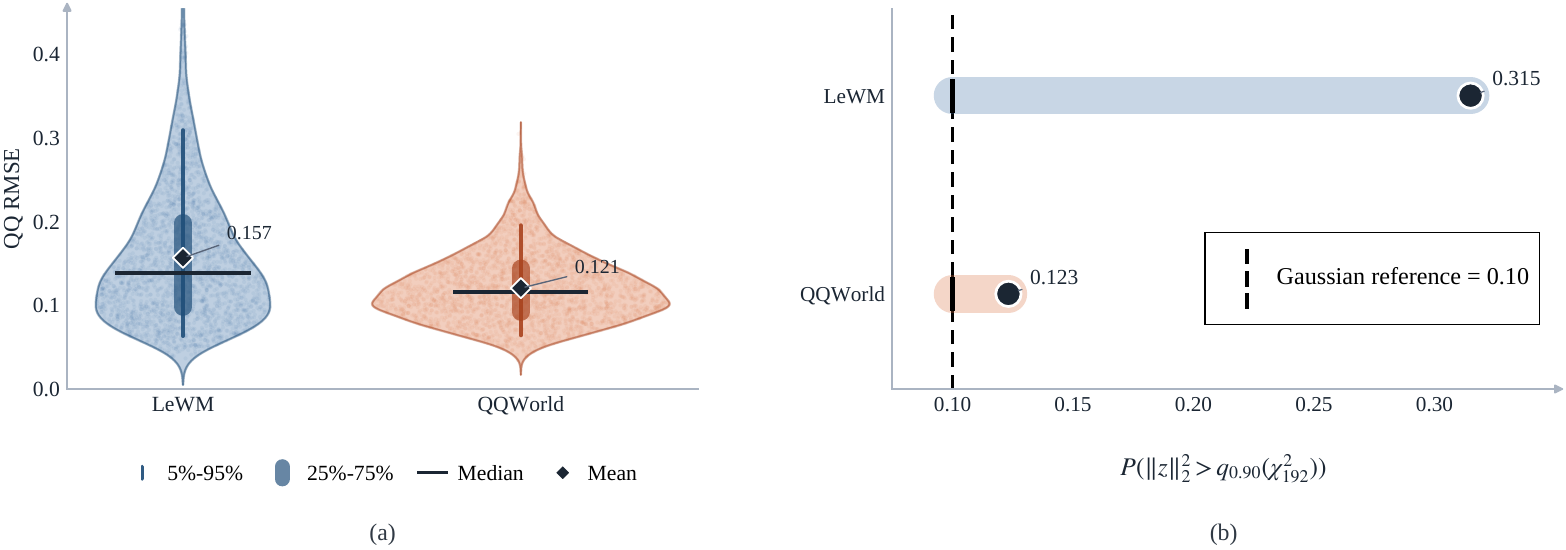}
\vspace{-1mm}
\caption{Analysis of the learned latent distributions. \textbf{Left}: QQ RMSE measures the discrepancy between the empirical quantiles of each one-dimensional latent projection and the corresponding standard Gaussian quantiles. Each dot represents one projection, while the violin width reflects the density of QQ RMSE values across projections. Compared with LeWM, QQWorld shifts the distribution toward smaller errors and reduces the mean QQ RMSE from \(0.157\) to \(0.121\), indicating closer agreement with the isotropic Gaussian target. \textbf{Right}: Radial tail comparison. For a sample drawn from a 192-dimensional standard Gaussian distribution, its squared Euclidean norm  follows a chi-squared distribution \(\chi^2_{192}\). We therefore report the probability $P\!\left(\lVert z\rVert_2^2 >q_{0.90}(\chi^2_{192})\right)$, which describes the portion of samples with squared norm larger than \(q_{0.90}(\chi^2_{192})\), i.e., the 90th percentile of the chi-squared distribution. Larger probability indicates heavier radial tails. QQWorld reduces the tail rate from \(0.315\) to \(0.123\), substantially approaching the Gaussian reference.
}
\vspace{-3mm}
\label{fig:tworoom_comparison}
\end{figure*}

Despite this regularization, we observe that the latents learned by LeWM have pronounced heavy tails, as shown in Figure~\ref{fig:tworoom_comparison} (Right). Such heavy-tailed behavior is undesirable in a latent world model: extreme latent values can push the learned dynamics into poorly represented regions, potentially amplifying errors during multi-step rollouts, while increasing the mismatch between the learned latent distribution and the target Gaussian prior. Crucially, this happens even though the model is explicitly penalized for non-normality, which suggests that the limitation lies in the optimization geometry of the particular test being used.

We analyze this limitation in two steps. First, we use the equivalence between the EP statistic and the squared Maximum Mean Discrepancy (MMD) to reinterpret the EP penalty as a unit-bandwidth kernel discrepancy from \(\mathcal{N}(0,1)\). Second, we analyze the gradient induced by this objective and show that its corrective force decays rapidly for latent values far from the bulk. Consequently, once a latent coordinate moves beyond the interaction scale of the kernel, the EP regularizer provides little restoring force and therefore fails to effectively suppress heavy-tailed deviations. 

To address this limitation, we replace the EP penalty with a Quantile–Quantile (QQ) matching objective that aligns the ordered latent values with the corresponding Gaussian quantiles. We refer to the resulting world model as \emph{QQWorld}. This objective provides a direct rank-matched transport signal whose magnitude grows linearly with the quantile deviation, rather than vanishing for extreme latent values. It therefore acts strongly and symmetrically on deviations in both tails. As shown in Figure~\ref{fig:tworoom_comparison}, QQWorld suppresses heavy tails and substantially improves the alignment of the learned latents with the target Gaussian distribution. Extensive experiments in Section~\ref{sec:experiments} also demonstrate that QQWorld consistently improves downstream planning performance.

Our contributions are summarized as follows.
\begin{itemize}
    \item We identify a key limitation of the EP regularizer: its corrective force rapidly vanishes for tail samples, leaving heavy-tailed deviations insufficiently controlled.
    \item We propose {QQWorld}, which replaces EP regularization with a quantile--quantile matching objective that maintains effective corrective gradients for tail deviations. QQWorld is a drop-in replacement for LeWM and introduces no additional hyperparameters.
    \item We develop a memory-efficient cross-batch strategy that enlarges the effective ranking pool without incurring significant GPU memory overhead. We also characterize its bias--variance trade-off.
\end{itemize}

\section{Related Work}
\subsection{Latent World Models for Planning}
World models learn predictive dynamics that enable an agent to evaluate the consequences of candidate actions before interacting with the environment. Early visual world models learn to reconstruct or generate future observations in pixel space~\citep{ha2018worldmodels,hafner2019planet,hafner2020dreamer}. Although pixel prediction provides rich supervision, it requires modeling high-dimensional observations and low-level visual details that may be irrelevant to control, making repeated model rollouts expensive during planning.

Joint-Embedding Predictive Architecture (JEPA) provides a reconstruction-free alternative by predicting future representations directly in latent space~\citep{lecun2022path,assran2025v,cui2026generalization}. This paradigm has recently been adopted for visual planning, such as \citet{3780338.3783523,NEURIPS2025_3e7cf447,fastlewm2026,nguyen2026latent,gao2026imwm,zhang2026hierarchical,josephinterpreting,zhang2026geoworld,masip2026ff,chen2026lawam}. Whereas these works primarily focus on world-model architectures and planning mechanisms, our work studies how the latent distribution itself should be regularized.

\subsection{Latent Regularization in World Models}
A central challenge in latent world modeling is representation collapse, where the encoder maps different observations to identical or weakly varying representations. \citet{bardes2022vicreg} propose to prevent collapse by explicitly controlling the variance of the learned embeddings. Such moment-based regularization is simple and effective, but constrains only low-order statistics and does not determine the full latent distribution. To address this issue, LeWM~\citep{leworldmodel2026} introduces EP test to encourage the full latent distribution to stay close to Gaussian distribution. Subsequent methods modify the structure imposed on the latent space. Sub-JEPA~\citep{zhao2026subjepasubspacegaussianregularization} applies Gaussian regularization within multiple subspaces rather than directly in the full ambient space. SD-JEPA~\citep{thil2026subspacedecomposedjepasdisentanglingprogression} decomposes the latent representation into progression and content subspaces, while SMWM~\citep{ivashkov2026sensorimotorworldmodelsperception} incorporates inverse-dynamics supervision.

These methods improve latent learning through subspace structures or auxiliary objectives. In contrast, we revisit the distribution-matching objective itself. We show that the corrective gradients of EP rapidly vanish for tail samples, while QQWorld effectively addresses this limitation by matching projected samples to their rank-aligned Gaussian quantiles.

\subsection{Statistical Tests and Distribution Matching}
Goodness-of-fit and two-sample testing are classical statistical problems for assessing whether observed samples follow a reference distribution or whether two samples share the same distribution~\citep{smirnov1948table,epps1983test,gretton2012kernel,hu2024two}. Classical test statistics include Kolmogorov--Smirnov (KS), EP~\citep{epps1983test}, and MMD~\citep{gretton2012kernel}. As shown in Section~\ref{sec:method}, the EP objective can be equivalently interpreted as an MMD with a Gaussian kernel and a fixed Gaussian reference distribution.

An alternative class of discrepancies is based on Wasserstein distance~\citep{villani2009optimal,hu2025two}. Sliced Wasserstein distances extend basic Wasserstein test to high dimensions through one-dimensional projections~\citep{nietert2022statistical}. In one dimension, the squared 2-Wasserstein distance is the squared L2 distance between quantile functions, linking its empirical form to quantile--quantile comparisons~\citep{shapiro1965analysis,wilk1968,ramdas2017wasserstein}. Despite their longstanding use for statistical diagnostics and distribution comparison, QQ-based objectives have rarely been explored for regularizing latent world models. Concurrent with our work, \citet{visreg2026} uses sliced Wasserstein matching for JEPA training, which focuses on self-supervised representation learning rather than world modeling and planning, and requires an additional variance term for scale control. In contrast, QQWorld directly replaces the EP regularizer in an end-to-end latent world model using a single quantile-matching objective and introduces no additional method-specific hyperparameters. We further introduce cross-batch QQ to enlarge the effective ranking pool without increasing memory cost, together with an analysis of its bias--variance trade-off.

\section{Method}\label{sec:method}
Let $z_1,\dots,z_N \in \mathbb{R}^d$ denote the latent embeddings of a training batch. Similar to~\citet{balestriero2025lejepa}, normality is enforced along random one-dimensional projections. For a direction $u\sim\operatorname{Uniform}(\mathbb{S}^{d-1})$ where $\mathbb{S}^{d-1}$ denotes the unit sphere in $\mathbb{R}^d$, we form the projected samples $x_n = \langle u, z_n\rangle$ and penalize their departure from $\mathcal{N}(0,1)$. Averaging over \(S\) independently sampled directions gives the regularizer:
\begin{equation}\label{eq:regularizer}
    \mathcal{R}(\{z_n\}) = \frac{1}{S}\sum_{s=1}^{S}\mathcal{L}\big(\{\langle u_s, z_n\rangle\}_{n=1}^N\big).
\end{equation}
By the Cram\'er--Wold theorem~\citep{cramer1936some}, a distribution on \(\mathbb{R}^d\) is determined by all of its one-dimensional projections. The finite-direction objective above can therefore be viewed as a Monte Carlo approximation to a sliced discrepancy that targets the joint latent distribution, rather than only its marginals. In the remainder of this section, we analyze a single slice and write $\mathcal{X}=\{x_n\}_{n=1}^N$ for the projected samples. Our contribution lies in the choice of the per-slice discrepancy $\mathcal{L}$ in Eq.~\ref{eq:regularizer}.

\subsection{Vanishing Gradients of EP Test}
As in~\citet{leworldmodel2026,fastlewm2026}, the world model is trained with the EP test statistic~\citep{epps1983test,balestriero2025lejepa}:
\begin{equation}\label{eq:EPtest}
    \mathcal{L}_{\mathrm{EP}}(\mathcal{X})=N \int_{-\infty}^\infty \big|\hat{\psi}(t) - \psi(t)\big|^2 w(t)\, dt,
\end{equation}
where $\hat{\psi}(t) = \frac{1}{N} \sum_{n=1}^N e^{itx_n}$ is the empirical characteristic function of $\mathcal{X}$, $\psi(t)$ is the characteristic function of $\mathcal{N}(0,1)$, and $w(t)=e^{-t^2/2}$ is the weighting function.

The EP statistic admits an equivalent kernel representation~\citep{rustamov2021closed}. In particular, up to a positive multiplicative constant, it is the squared MMD between the empirical latent distribution and \(\mathcal{N}(0,1)\), under the Gaussian kernel \(k(x,y)=e^{-{(x-y)^2}/{2}}.\) More precisely,
\begin{align}
    \mathcal{L}_\mathrm{EP} &= \sqrt{2\pi}\, N \cdot \mathrm{MMD}^2\big(\{x_n\}, \mathcal{N}(0,1)\big) \label{eq:MMD} = \frac{\sqrt{2\pi}}{N} \sum_{m,n} e^{-\frac{(x_m - x_n)^2}{2}} - {2\sqrt{\pi}} \sum_n e^{\frac{-x_n^2}{4}} + \sqrt{\tfrac{2\pi}{3}}\, N .
\end{align}
The Gaussian weighting $w(t)$ in Eq.~\ref{eq:EPtest} induces the Gaussian kernel $k(x,y)$ of the MMD, which imposes a limited interaction scale in the sample domain. 

We differentiate this objective with respect to a single latent coordinate:
\begin{equation}
    \frac{\partial \mathcal{L}_{\mathrm{EP}}}{\partial x_n}
    =
    \underbrace{
        -\frac{2\sqrt{2\pi}}{N}
        \sum_m (x_n-x_m)\,e^{-(x_n-x_m)^2/2}
    }_{\text{pairwise repulsive term}}
    +
    \underbrace{
        \vphantom{
            \frac{2\sqrt{2\pi}}{N}
            \sum_m (x_n-x_m)\,e^{-(x_n-x_m)^2/2}
        }
        \sqrt{\pi}\,x_n\,e^{-x_n^2/4}
    }_{\text{center-attraction term}}.
    \label{eq:ep-grad}
\end{equation}

The two components of Eq.~\ref{eq:ep-grad} play complementary roles. The pairwise term pushes $x_n$ away from other samples $x_m$ and thereby prevents representation collapse. The second term arises from the interaction between the empirical distribution and the Gaussian target. Its negative gradient always points toward the origin, encouraging the projected features to remain within the high-density region of the standard Gaussian. 

The following proposition characterizes this gradient for an outlier value that lies in the tail of the projected samples.

\begin{proposition}[Vanishing restoring force]
\label{prop:vanish}
Let $x_n = h > 0$ be a coordinate in the tail while the remaining $N-1$ samples lie in a bulk $\{|x_m|\le R\}$ for a fixed $R$. Then, as $h \to \infty$, the pairwise term of Eq.~\ref{eq:ep-grad} is $O\big(h\,e^{-(h-R)^2/2}\big)$ and is asymptotically negligible relative to the center-attraction term, so that the restoring force pulling $h$ back toward the bulk satisfies
\begin{equation}
    \frac{\partial \mathcal L_{\mathrm{EP}}}{\partial h}
    \;=\; \sqrt{\pi}\, h\, e^{-h^2/4}\big(1+o(1)\big).
\end{equation}
Moreover, the magnitude of this asymptotic gradient attains its maximum at $h=\sqrt{2}$ and decays super-exponentially thereafter.
\end{proposition}

As indicated by Proposition~\ref{prop:vanish}, for a coordinate that has escaped into the tail, the restoring force that would pull it back toward the bulk rapidly decays toward zero. In other words, once a latent leaves the interaction scale, the EP regularizer stops seeing it, and heavy tails are free to persist and grow, making the EP test poorly suited to suppressing heavy tails.

\subsection{QQ Regularization}
The above analysis suggests that an effective alternative to the EP regularizer should have meaningful corrective gradients for tail deviations. To this end, we define the quantile--quantile (QQ) matching loss as:
\begin{align}
    \mathcal{L}_\mathrm{QQ}(\mathcal{X}) = \sum_{n=1}^N (\hat{x}_n - q_n)^2, \quad q_n = \Phi^{-1}\!\Big(\tfrac{n-0.5}{N}\Big),
    \label{eq:qq}  
\end{align}
where $\hat{x}_1\le\cdots\le\hat{x}_N$ are the order statistics of the projected batch $\mathcal{X}$, and $q_n$ are the corresponding Gaussian quantiles. $\Phi^{-1}$ denotes the inverse cumulative distribution function of the standard Gaussian distribution. The resulting world model, trained with the QQ regularizer in Eq.~\ref{eq:qq}, is referred to as \emph{QQWorld}. This construction is motivated by the classical QQ-based distributional comparisons~\citep{shapiro1965analysis,wilk1968,ramdas2017wasserstein}.

The loss in Eq.~\ref{eq:qq} admits a natural optimal transport interpretation. In parallel to Eq.~\ref{eq:MMD}, our $\mathcal{L}_\mathrm{QQ}$ can be seen as a quadrature approximation of the squared $2$-Wasserstein distance to the target:
\begin{equation}
    \mathcal{L}_{\mathrm{QQ}}(\mathcal{X}) \;\approx\; N\cdot W_2^2\big(\{x_n\},\mathcal{N}(0,1)\big).
\end{equation}
Thus, whereas the EP regularizer measures distributional discrepancy through a limited-bandwidth kernel, the QQ regularizer directly matches empirical and Gaussian quantiles.

\begin{proposition}[Non-vanishing restoring force]
\label{prop:qq}
Let $\rho(n)$ denote the rank of $x_n$ within $\mathcal X$. The proposed regularization $\mathcal{L}_\mathrm{QQ}(\mathcal{X})$ can be equivalently written as:
\begin{equation} 
\sum_{n=1}^N (x_n - q_{\rho(n)})^2 = \sum_{n=1}^N \bigg[x_n - \Phi^{-1}\bigg(\frac{\rho(n) -0.5}{N}\bigg)\bigg]^2.
\label{eq:qq-equiv}
\end{equation}
Then $\mathcal{L}_{\mathrm{QQ}}$ is differentiable at every configuration without ties, with
\begin{equation}\label{eq:qq-grad}
    \frac{\partial \mathcal{L}_{\mathrm{QQ}}}{\partial x_n} = 2\big(x_n - q_{\rho(n)}\big).
\end{equation}
Consequently, the negative-gradient update acting on $x_n$ points directly toward the rank-matched Gaussian quantile and has magnitude $2|x_n-q_{\rho(n)}|$.
\end{proposition}

Proposition~\ref{prop:qq} shows that each projected latent $x_n$ receives a direct rank-matched transport signal toward its corresponding Gaussian quantile $q_{\rho(n)}$. Unlike the EP gradient in Eq.~\ref{eq:ep-grad}, which vanishes for extreme values, the QQ gradient becomes stronger as the quantile discrepancy increases $|x_n-q_{\rho(n)}|$, alleviating the problem of heavy tails. 

\subsubsection{Behavior Near Rank-Switching Boundaries}
Although the QQ objective Eq.~\ref{eq:qq-equiv} is continuous and differentiable almost everywhere, its gradient Eq.~\ref{eq:qq-grad} changes discontinuously when two projected samples exchange ranks, because their matched Gaussian quantiles are exchanged as well. For example, if the rank of a sample $x_n$ changes from $k$ to $k+1$, its gradient jumps from $2\big(x_n - q_{k}\big)$ to $2\big(x_n - q_{k+1}\big)$. At first sight, such rank-induced gradient discontinuity may appear to hinder optimization. We show, however, that it is not a problem in practice: a tied configuration is locally repelling under the QQ objective, and the loss encourages the two samples to separate.

To see this, consider two samples associated with adjacent ranks $k$ and $k+1$. Around a common center $c$, parameterize them as
\begin{equation}
    \hat{x}_k=c-\delta,
    \qquad
    \hat{x}_{k+1}=c+\delta,
    \qquad
    \delta\geq0,
    \label{eq:qq-tie-perturbation}
\end{equation}
and a rank exchange can occur only when they meet at a tie, corresponding to $\delta \rightarrow 0$. Their contribution to the QQ loss is
\begin{equation}
    \ell_k(\delta)
    =
    \left(c-\delta-q_k\right)^2
    +
    \left(c+\delta-q_{k+1}\right)^2.
    \label{eq:qq-tie-loss}
\end{equation}
Differentiating with respect to the separation parameter gives
\begin{equation}
    \frac{d\ell_k(\delta)}{d\delta}
    =
    4\delta
    -
    2\left(q_{k+1}-q_k\right).
    \label{eq:qq-tie-derivative}
\end{equation}
Since $q_{k+1}>q_k$, the one-sided directional derivative at the tied configuration is strictly negative:
\begin{equation}
    \left.
    \frac{d\ell_k(\delta)}{d\delta}
    \right|_{\delta=0^+}
    =
    -2\left(q_{k+1}-q_k\right)
    <0.
    \label{eq:qq-tie-direction}
\end{equation}
Therefore, the gradient descent step of $\mathcal{L}_{QQ}$ will enlarge $\delta$ and increase the separation between $\hat{x}_k$ and $\hat{x}_{k+1}$. In other words, the tie boundary is repelling rather than attractive. Samples assigned to different ranks are pulled toward distinct Gaussian quantiles and are therefore encouraged to move away from the nondifferentiable configuration. This local anti-collapse behavior explains why the apparent gradient discontinuity caused by sorting does not create difficulty in practice.

\subsubsection{Relation between QQ and EP}
\label{subsec:qq-ep-relation}
We next establish a one-way relation between the QQ and EP objectives. Specifically, driving the QQ loss to zero also drives the EP loss to zero. The converse, however, does not hold: the Gaussian kernel underlying EP can assign an asymptotically negligible penalty to a fraction of arbitrarily distant tail observations, whereas QQ directly penalizes their squared quantile deviations.

\begin{proposition}[One-way control between QQ and EP]
\label{prop:qq-ep-control}
For the EP loss in Eq.~\ref{eq:EPtest}, there exists a constant $C>0$, independent of $\mathcal{X}$ and $N$, such that
\begin{equation}
    \mathcal{L}_{\mathrm{EP}}(\mathcal{X})
    \leq
    C\left\{
        \mathcal{L}_{\mathrm{QQ}}(\mathcal{X})
        +
        \frac{\log N}{N}
    \right\}.
    \label{eq:qq-controls-ep}
\end{equation}
Consequently, for a large $N$,
\begin{equation}
    \mathcal{L}_{\mathrm{QQ}}(\mathcal{X})\to0
    \quad\Longrightarrow\quad
    \mathcal{L}_{\mathrm{EP}}(\mathcal{X})\to0.
    \label{eq:qq-implies-ep}
\end{equation}

The converse does not hold. In particular, it is possible that
\begin{equation}
    \mathcal{L}_{\mathrm{EP}}(\mathcal{X})\to0,
    \qquad\text{while}\qquad
    \mathcal{L}_{\mathrm{QQ}}(\mathcal{X})\to\infty.
    \label{eq:ep-not-imply-qq}
\end{equation}
\end{proposition}
Proposition~\ref{prop:qq-ep-control} shows that QQ matching provides a stronger form of distributional control than EP matching: driving the empirical order statistics toward their corresponding Gaussian quantiles also drives the EP discrepancy to zero. The converse does not hold: a small EP loss may coexist with arbitrarily large deviations in tail observations, because their Gaussian-kernel contributions in Eq.~\ref{eq:MMD} saturate with distance. This one-way control further supports the use of QQ regularization in world modeling. 

\subsection{Cross-Batch QQ}
\label{subsec:qq-queue}
As shown in Eq.~\ref{eq:qq-equiv}, the accuracy of empirical QQ matching depends on the number of samples used to estimate the ranks. A larger batch provides more accurate Gaussian quantile targets, but also requires storing more intermediate activations for backpropagation. To enable lightweight training under a limited memory budget, we introduce \emph{Cross-Batch QQ}, which enlarges the ranking set using detached features from recent iterations while keeping the backpropagation batch size unchanged.

Let $x_1^{(t)},\ldots,x_N^{(t)}$ denote the projected features in the current iteration $t$, and let a first-in--first-out queue retain the projected features from the previous $K$ iterations. The pooled ranking set is
\begin{equation}
\mathcal{P}_{t,K}
=
\bigcup_{j=0}^{K}
\{x_1^{(t-j)},\ldots,x_N^{(t-j)}\},
~~
M=(K+1)N, 
\label{eq:qq-pool}
\end{equation}
where the historical features with $j\geq1$ are detached from the computation graph. Hence, $M$ samples participate in rank estimation, whereas gradients are propagated through only the $N$ current features $\{x_1^{(t)},\ldots,x_N^{(t)}\}$.

For a current feature $x_n^{(t)}$, let $\rho_{t,K}(n)\in\{1,\ldots,M\}$ denote its rank within $\mathcal{P}_{t,K}$. Based on Eq.~\ref{eq:qq-equiv}, the queue-based objective is
\begin{equation}
\mathcal{L}_{\mathrm{QQ}}^{\mathrm{queue}}
=
\sum_{n=1}^{N}
\left[
x_n^{(t)}
-
\Phi^{-1}\!\left(
\frac{\rho_{t,K}(n)-0.5}{M}
\right)
\right]^2.
\label{eq:queue-qq}
\end{equation}
Because the historical features are used only for ranking, Cross-Batch QQ increases the effective ranking-set size from $N$ to $M$ without increasing the number of samples whose computation graphs must be retained. It therefore obtains much of the rank-estimation benefit of a larger batch at substantially lower memory cost.

\subsubsection{Bias--Variance Tradeoff}
To understand the statistical effect of the queue, let $F_t$ denote the distribution of projected latent features produced by the encoder at iteration $t$. At the population level, a feature $x$ should ideally be matched to
\begin{equation}
    q(x)
    =
    \Phi^{-1}\!\bigl(F_t(x)\bigr),
\end{equation}
which is its corresponding standard Gaussian quantile. In practice, $F_t(x)$ is unknown and is replaced by an empirical estimate computed from the pooled ranking set: $\widehat{F}_{t,K}(x_n) = \frac{\rho_{t,K}(n)-0.5}{M}$. Thus, $\hat q(x) = \Phi^{-1}(\widehat{F}_{t,K}(x))$ is the empirical learning target used by Cross-Batch QQ, whereas $q(x)$ is the ideal population target.

The queue strategy introduces a bias--variance trade-off in estimating this ideal target. Since $\mathcal P_{t,K}$ contains features produced at iterations $t,t-1,\ldots,t-K$, its empirical CDF $\widehat F_{t,K}$ is an estimation of the mixture distribution
\begin{equation}
    \bar F_{t,K}
    =
    \frac{1}{K+1}
    \sum_{j=0}^{K}F_{t-j},
    \label{eq:temporal-mixture}
\end{equation}
rather than the current distribution $F_t$. If the representation changes negligibly over the queue window, then $F_{t-j}\approx F_t$ and increasing $K$ mainly reduces the sampling variance of the empirical estimate. During training, however, encoder updates generally yield $F_{t-j}\neq F_t$, so the variance reduction is accompanied by representation-staleness bias.

To understand the bias-variance trade-off, we quantify the quality of the learning target $\hat{q}(x)$ by its mean squared error relative to the ideal target $q(x)$:
\begin{align} 
\operatorname{MSE}\bigl(\hat q(x)\bigr) = \mathbb{E}[\hat q(x) - q(x)]^2 \label{eq:queue-bias-variance}
\approx
\frac{1}{
\phi(q(x))^2
}
\bigg[
\underbrace{
\frac{F_t(x)(1-F_t(x))}{N(K+1)}
}_{\text{rank-estimation variance}}
+  \underbrace{
\vphantom{
\frac{F_t(x)(1-F_t(x))}{N(K+1)}
}
\bigl\{
\bar{F}_{t,K}(x)-F_t(x)
\bigr\}^2
}_{\text{representation-staleness bias}}
\bigg], 
\end{align}
where the expectation is over the pooled samples used for estimation, and $\phi$ is the standard Gaussian density. 

For comparison, the standard QQ in Eq.~\ref{eq:qq} corresponds to $K=0$ in Eq.~\ref{eq:queue-bias-variance} and has the approximate target-estimation error:
\begin{equation}
\frac{1}{
\phi(q(x))^2
}  \frac{F_t(x)(1-F_t(x))}{N}.
\label{eq:bias-variance}
\end{equation}
From Eq. \ref{eq:queue-bias-variance} and Eq. \ref{eq:bias-variance}, increasing $K$ reduces the variance term from order $N^{-1}$ to order $\{N(K+1)\}^{-1}$, but may introduce a nonzero staleness bias.

Therefore, Cross-Batch QQ is most useful in the small-batch regime (small $N$), where the variance of current-batch ranks is substantial and a short queue can provide more reliable learning targets. When the current batch is already sufficiently large, the additional variance reduction becomes limited and may be outweighed by representation staleness. We consequently use Cross-Batch QQ as an optional memory-efficient training strategy: it enables a small backpropagation batch to recover much of the rank-estimation benefit of a larger batch, while our base QQ loss is still preferable when memory permits a sufficiently large batch.

\begin{table*}[t]
\centering
\small
\renewcommand{\arraystretch}{1.08}
\setlength{\tabcolsep}{4pt}
\begin{tabular*}{\textwidth}{ @{\extracolsep{\fill}} l c c c c c @{} }
\toprule
\textbf{Method}
& \textbf{Two-Room}
& \textbf{Reacher}
& \textbf{PushT}
& \textbf{OGBench-Cube}
& \textbf{Avg.} \\
\midrule
PLDM \citep{NEURIPS2025_3e7cf447}
& 97.00
& 78.00
& 78.00
& 65.00
& 79.50 \\
DINO-WM (w/o proprio.) \citep{3780338.3783523}
& 100.00
& 79.00
& 74.00
& 86.00
& 84.75 \\
DINO-WM (w/ proprio.) \citep{3780338.3783523}
& 100.00
& \textemdash
& 92.00
& \textemdash
& \textemdash \\
\midrule
LeWM \citep{leworldmodel2026}
& \(84.33 {\scriptstyle\pm 4.23}\)
& \(82.67 {\scriptstyle\pm 4.42}\)
& \(84.67 {\scriptstyle\pm 6.53}\)
& \(67.33 {\scriptstyle\pm 5.01}\)
& 79.75 \\
Sub-JEPA \citep{zhao2026subjepasubspacegaussianregularization}
& \(\mathbf{93.67} {\scriptstyle\pm 4.27}\)
& \(81.00 {\scriptstyle\pm 2.10}\)
& \(89.00 {\scriptstyle\pm 5.33}\)
& \(69.00 {\scriptstyle\pm 8.69}\)
& \underline{83.17} \\
SD-JEPA \citep{thil2026subspacedecomposedjepasdisentanglingprogression}
& \(86.33 {\scriptstyle\pm 6.12}\)
& \(\underline{85.00} {\scriptstyle\pm 5.02}\)
& \(\underline{89.67} {\scriptstyle\pm 4.27}\)
& \(69.67 {\scriptstyle\pm 5.13}\)
& 82.67 \\
SMWM \citep{ivashkov2026sensorimotorworldmodelsperception}
& \(\underline{88.67} {\scriptstyle\pm 7.12}\)
& \(73.00 {\scriptstyle\pm 4.69}\)
& \(86.00 {\scriptstyle\pm 2.53}\)
& \(\mathbf{84.33} {\scriptstyle\pm 4.97}\)
& 83.00 \\
\midrule
\method{} (Ours)
& \(\mathbf{93.67} {\scriptstyle\pm 3.44}\)
& \(\mathbf{85.33} {\scriptstyle\pm 5.16}\)
& \(\mathbf{91.00} {\scriptstyle\pm 5.76}\)
& \(\underline{70.33} {\scriptstyle\pm 7.31}\)
& \(\mathbf{85.08}\) \\
\bottomrule
\end{tabular*}
\caption{Planning success rate (\%, higher is better) across four environments. Values are reported as mean \(\pm\) standard deviation across six random seeds. 
}
\label{tab:planning_success_rates}
\end{table*}

\section{Experiments}
\label{sec:experiments}

\subsection{Experimental Setup}
We follow LeWM~\citep{leworldmodel2026} for dataset preprocessing, model training, and CEM-based goal-conditioned evaluation. Experiments are conducted on the same offline datasets, including Two-Room~\citep{NEURIPS2025_3e7cf447}, PushT~\citep{3780338.3783523}, Reacher~\citep{tassa2018deepmindcontrolsuite}, and OGBench-Cube~\citep{park2025ogbench}. QQWorld differs from LeWM only by replacing the EP regularizer with the proposed QQ objective. We set the QQ regularization weight to $3.5$ for all environments. Unlike baseline methods~\citep{thil2026subspacedecomposedjepasdisentanglingprogression,zhao2026subjepasubspacegaussianregularization} which introduce additional hyperparameters that need to be tuned, QQWorld does not have new hyperparameters or require extra tuning. 

\subsection{Planning Performance}
We compare QQWorld with LeWM~\citep{leworldmodel2026} and its recent variants, including Sub-JEPA~\citep{zhao2026subjepasubspacegaussianregularization}, SD-JEPA~\citep{thil2026subspacedecomposedjepasdisentanglingprogression}, and SMWM~\citep{ivashkov2026sensorimotorworldmodelsperception}. We use their official implementations and evaluate all methods under the same planning protocol with the six random seeds adopted by Sub-JEPA. For baselines that introduce method-specific hyperparameters, we use their best-performing hyperparameters which achieve the highest average success rate across the four environments. Since official checkpoints for PLDM~\citep{NEURIPS2025_3e7cf447} and DINO-WM~\citep{3780338.3783523} are not available, we cite their success rates directly from LeWM under the same evaluation protocol.

As shown in Table~\ref{tab:planning_success_rates}, QQWorld achieves the highest average success rate of \(85.08\%\), improving the LeWM baseline by \(5.33\) percentage points. Moreover, it consistently outperforms LeWM on all four environments, indicating that the better tail correction provided by QQ regularization translates into improved downstream planning performance.

\begin{table}[t]
\centering
\setlength{\tabcolsep}{25pt}
\renewcommand{\arraystretch}{1.12}
\begin{tabular}{@{}lcc@{}}
    \toprule
    Method
    & KS \(\downarrow\)
    & EP \(\downarrow\) \\
    \midrule
    LeWM~\citep{leworldmodel2026}
    & 0.038
    & 119.909 \\
    QQWorld
    & \textbf{0.032}
    & \textbf{82.294} \\
    \midrule
    \text{Relative reduction (\%)}
    & 15.8
    & 31.4 \\
    \bottomrule
\end{tabular}
\caption{Normality comparison between LeWM and QQWorld. Results are averaged over four environments. KS and EP denotes the Kolmogorov--Smirnov and Epps--Pulley statistics, respectively. Lower values indicate closer agreement with the standard Gaussian distribution.
}
\label{tab:overall_gaussian_assessment}
\end{table}

\begin{figure}[t]
\centering
\includegraphics[width=0.55\textwidth]{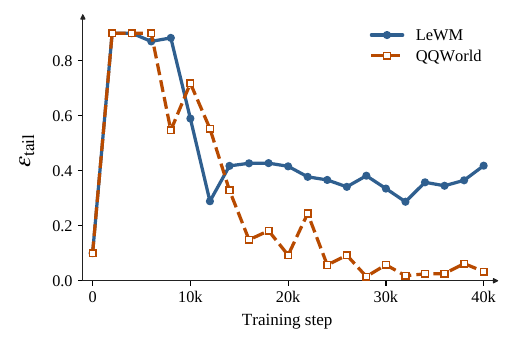}
\caption{Evolution of the tail over 40k training steps on Two-Room. We report \(\varepsilon_{\mathrm{tail}}=\left|{P}\!\left(\lVert z\rVert_2^2>q_{0.90}(\chi^2_{192})\right)-0.10\right|\), where the probability $P$ represents the proportion of latent samples exceeding the $90$th percentile of $\chi^2_{192}$. The Gaussian reference value is $0.10$. Lower calibration error $\varepsilon_\text{tail}$ indicates closer agreement with the standard Gaussian tail.
}
\label{fig:two_room_training_test}
\end{figure}

\subsection{Normality Assessment}
To verify the effectiveness of the proposed QQ loss, we assess the Gaussianity of the encoder latents learned by LeWM and QQWorld under the same evaluation protocol. For each method, we randomly sample 20,000 latents and project them along 6,144 random unit directions. The sampled latents and projection directions are fixed across all comparisons.

Figure~\ref{fig:tworoom_comparison} shows that QQWorld achieves a lower QQ RMSE and substantially improves the alignment of the latents with the standard Gaussian distribution. To provide complementary normality measures, we also report the KS and EP statistics in Table~\ref{tab:overall_gaussian_assessment}. QQWorld reduces the KS statistic by 15.8\% and the EP statistic by 31.4\% relative to LeWM, indicating consistently better Gaussian alignment. Notably, QQWorld achieves a lower EP statistic even though LeWM is trained directly with the EP objective, further demonstrating the effectiveness of the proposed QQ regularization.

We further examine how the latent tails evolve during training. As shown in Figure~\ref{fig:two_room_training_test}, QQWorld progressively suppresses the heavy radial tails, with the tail probability steadily decreasing toward the standard Gaussian reference. In contrast, LeWM retains substantially heavier tails throughout training. This result confirms the strong capability of the proposed QQ regularization in correcting tail deviations.

\begin{table}[t]
\centering
\setlength{\tabcolsep}{15.pt}
\renewcommand{\arraystretch}{1.05}
\begin{tabular}{@{}llcccc@{}}
\toprule
Property & Model
& \multicolumn{2}{c}{Linear}
& \multicolumn{2}{c}{MLP} \\
\cmidrule(lr){3-4}
\cmidrule(lr){5-6}
& & MSE $\downarrow$ & $r\uparrow$
& MSE $\downarrow$ & $r\uparrow$ \\
\midrule

Agent location
& LeWM
& 0.042 & \textbf{0.979}
& \textbf{0.001} & 0.999 \\
& \method{} 
& \textbf{0.041} & \textbf{0.979}
& \textbf{0.001} & \textbf{1.000} \\

\midrule
Block location
& LeWM
& 0.022 & 0.989
& \textbf{0.000} & \textbf{1.000} \\
& \method{}
& \textbf{0.021} & \textbf{0.990}
& \textbf{0.000} & \textbf{1.000} \\

\midrule
Block angle
& LeWM
& 0.176 & 0.908
& 0.008 & \textbf{0.996} \\
& \method{} 
& \textbf{0.172} & \textbf{0.910}
& \textbf{0.007} & \textbf{0.996} \\

\bottomrule
\end{tabular}
\caption{
Linear- and MLP-probing results for agent location, block location, and block angle. We report the mean squared error (MSE) and Pearson correlation coefficient \(r\). Lower MSE and higher \(r\) indicate better performance. The best results are shown in bold.
}
\label{tab:pusht_probe}
\end{table}

\subsection{Physical State Probing}
Planning performance measures how effectively a learned world model supports goal-directed control, but does not directly reveal what physical information is encoded in its latent representations. Therefore, similar to LeWM~\citep{leworldmodel2026}, we evaluate how accurately task-relevant physical states can be decoded from the learned latents on PushT.

Specifically, we probe three state properties: the agent location, block location, and block angle. For each property, we train both a linear predictor and a lightweight MLP on top of the frozen latent representations. All results are averaged over six random seeds. As shown in Table~\ref{tab:pusht_probe}, QQWorld consistently matches or outperforms LeWM across the probing tasks, indicating that QQ regularization effectively preserves task-relevant physical information in the learned latent space.

\subsection{Visualizing the Learned Latents}
\begin{figure}[tbp]
\centering
\paperfig[width=0.9\linewidth]{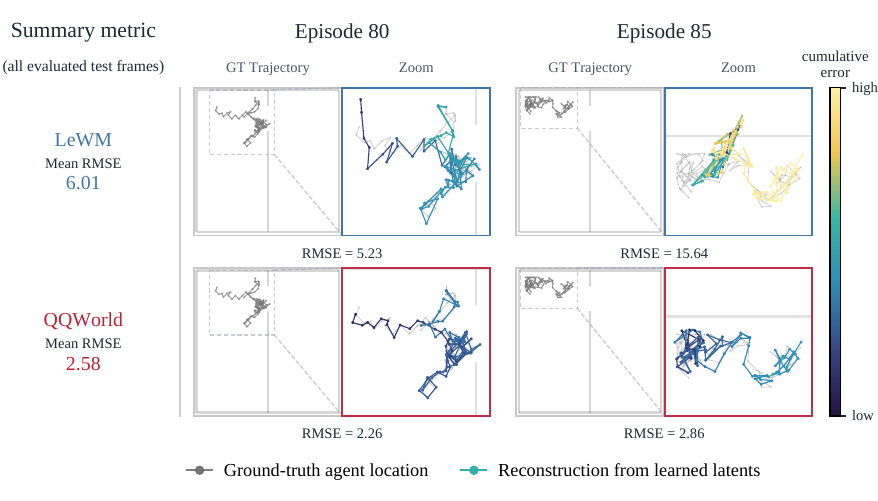}
\caption{
Visualization of the learned latent representations on Two-Room. We train a linear probe to predict the two-dimensional agent location from a latent representation. The figure visualizes two representative test episodes. For each episode, the left subpanel shows the ground-truth trajectory, while the right subpanel magnifies the region enclosed by the dashed box. Gray curves denote the ground-truth trajectories, and colored curves denote the trajectories decoded from the learned latents. The color along each predicted trajectory represents the cumulative RMSE up to the corresponding time step. The value below each example reports its episode RMSE, while the leftmost summary column reports the RMSE computed over all test episodes. 
}
\label{fig:tworoom_linear_probe}
\end{figure}

We further visualize the learned representations on Two-Room, which can be naturally described by the agent's two-dimensional location. We fit a linear readout that maps the frozen latent representation \(z_t\) to the ground-truth agent position \(p_t \in \mathbb{R}^2\). We train it on episodes \(0\)--\(79\) and evaluate on the held-out episodes \(80\)--\(99\). For an episode containing \(T\) frames, let \(\{\hat{p}_t\}_{t=1}^{T}\) and \(\{p_t\}_{t=1}^{T}\) denote the decoded and ground-truth trajectories, respectively. Their discrepancy is measured by \(\mathrm{RMSE}(T)=\sqrt{\sum_{t=1}^{T}\left\|\hat{p}_t-p_t\right\|_2^2/T
}.\) We additionally visualize the cumulative $\mathrm{RMSE}(t)$ at each time step \(t\) through the color of the predicted trajectory in Figure~\ref{fig:tworoom_linear_probe}. QQWorld produces decoded trajectories that more closely resemble the ground-truth paths, which accumulate less prediction error over time. These results further illustrate the effectiveness of QQ regularization in world modeling.

\begin{table}[t]
\centering
\setlength{\tabcolsep}{8pt}
\renewcommand{\arraystretch}{1.08}
\begin{tabular}{ccccc}
\toprule
Batch size 
& Queue length 
& Total size 
& Avg. success 
& Memory\\
\midrule
32  & 1 & 32  & 65.42 & 3434.7 \\
32  & 2 & 64  & 80.67 & 3434.9 \\
32  & 3 & 96  & \underline{83.50} & 3435.0 \\
32  & 4 & 128 & 81.75 & 3435.1 \\
\midrule
64  & 1 & 64  & 81.83 & 6552.9 \\
64  & 2 & 128 & 81.00 & 6553.1 \\
\midrule
128 & 1 & 128 & \textbf{85.08} & 12747.1 \\
128 & 2 & 256 & 80.83 & 12747.8 \\
\bottomrule
\end{tabular}
\caption{
Effect of the batch size $N$ and cross-batch queue length $K+1$ on planning success rate (\%) and GPU memory usage (MB). The total queue size is \(M=N(K+1)\) as defined in Eq.~\ref{eq:qq-pool}.
}
\label{tab:cross_batch_queue}
\end{table}

\subsection{Cross-Batch QQ}
The proposed QQ matching benefits from a large ranking pool. However, increasing the training batch size requires retaining activations for more samples and therefore incurs additional GPU memory overhead. To support memory-efficient training, we introduce cross-batch QQ, which augments the current batch with detached latent samples from previous batches. Quantile ranks are computed over all \(M=N(K+1)\) samples in the pooled set, whereas gradients are propagated only through the \(N\) samples in the current batch. Thus, cross-batch QQ decouples the effective ranking-pool size \(M\) from the backpropagation batch size \(N\).

Table~\ref{tab:cross_batch_queue} reports planning performance and GPU memory usage under different batch sizes and queue lengths. With \(N=32\), using only the current batch yields an average success rate of \(65.42\%\). Adding one and two historical batches increases the success rate to \(80.67\%\) and \(83.50\%\), respectively, while leaving GPU memory essentially unchanged. In particular, the configuration \(N=32,\ K+1=3\) outperforms LeWM (\(83.50\%\) versus \(79.75\%\)), although LeWM uses a batch size of \(128\) to obtain an accurate estimate of the characteristic function. Its performance is also close to the large-batch QQWorld result of \(85.08\%\), while reducing the training batch size by \(4\times\) and GPU memory usage by \(73\%\).

The benefit of a larger queue is not monotonic. Increasing the queue length from three to four batches at \(N=32\) reduces the success rate from \(83.50\%\) to \(81.75\%\). Similarly, adding a historical batch does not improve performance for \(N=64\) or \(N=128\). These observations are consistent with the bias--variance trade-off analyzed in Eq.~\ref{eq:queue-bias-variance}: a moderate queue reduces the variance of empirical quantile estimation when the current batch is small, whereas an excessively long queue introduces staleness bias because historical samples may no longer represent the current latent distribution.

Although a batch size of \(128\) appears affordable for the current model, cross-batch QQ provides a more scalable training mechanism by decoupling the ranking-pool size from the backpropagation batch size. This property becomes particularly relevant when scaling to larger world models, higher-resolution observations, or longer sequences, for which increasing the physical batch size can incur substantially greater memory overhead. Overall, cross-batch QQ recovers most of the performance benefit of large-batch quantile matching while substantially reducing GPU memory consumption.

\section{Conclusion}
We introduce QQWorld, a simple replacement for the EP regularizer in latent world models. By directly matching projected latent samples to Gaussian quantiles, QQWorld provides stronger correction for tail deviations. We further propose cross-batch QQ to improve quantile estimation under limited GPU memory. Extensive experiments show that QQWorld improves both latent Gaussianity and downstream planning performance. More broadly, our results highlight that a statistic effective for measuring distributional discrepancy is not necessarily an effective training objective for world models: beyond distinguishing distributions, the objective must also provide informative and well-behaved gradients throughout optimization.

\bibliographystyle{iclr2025_conference}
\bibliography{main}

\end{document}